\documentclass{article}
\PassOptionsToPackage{numbers, compress}{natbib}

\usepackage[preprint]{neurips_2026}
\usepackage{graphicx}
\usepackage{amsmath}
\usepackage{multirow} 

\usepackage[utf8]{inputenc} 
\usepackage[T1]{fontenc}    
\usepackage{hyperref}       
\usepackage{url}            
\usepackage{booktabs}       
\usepackage{amsfonts}       
\usepackage{nicefrac}       
\usepackage{microtype}      
\usepackage[table]{xcolor}
\usepackage{makecell}
\usepackage{array}

\definecolor{frblue}{HTML}{E8EAFF}
\definecolor{nrgreen}{HTML}{E8F8E8}
\definecolor{nonoptyellow}{HTML}{FFF9DF}
\definecolor{fidgray}{HTML}{F2F2F2}

\usepackage{booktabs}
\usepackage{multirow}
\usepackage{graphicx}

\definecolor{trainfr}{RGB}{225,238,255}      
\definecolor{trainnr}{RGB}{225,245,225}      
\definecolor{nonoptfr}{RGB}{240,230,255}     
\definecolor{nonoptnr}{RGB}{255,242,220}     
\definecolor{distgray}{RGB}{235,235,235}     

\newcommand{\redbest}[1]{\textbf{\textcolor{red}{#1}}}

\usepackage{tabularx}
\usepackage{placeins}
\usepackage{float}
\title{FoA-SR: Faithful or Aesthetic? Profile-Aware Preference Optimization for Real-World Image Super-Resolution}

\author{
  Amjad Mahdi Alqarni \\
  Department of Computer Science \\
  University of Kentucky \\
  Lexington, KY, USA \\
  \texttt{amjad.alqarni@uky.edu}
  \And
  Peizhong Ju \\
  Department of Computer Science \\
  University of Kentucky \\
  Lexington, KY, USA \\
  \texttt{peizhong.ju@uky.edu}
}

\begin{document}

\maketitle

\begin{abstract}
Real-world image super-resolution (SR) is often designed with a single restoration objective, despite the current capacity of generative models to produce multiple high-quality reconstructions for the same input. In this paper, we argue that the best restoration strategy is subject to the specific restoration profile: a Faithful restoration prioritizes reference consistency, structure preservation, and hallucination suppression, whereas an Aesthetic restoration prioritizes visually pleasing and natural-looking details. We propose \textbf{FoA-SR}, a novel preference optimization approach to real-world SR based on profiles. To achieve this goal, FoA-SR starts with our supervised FLUX.2-based SR adapter (Flux2SR) trained with LR latent conditioning, flow matching, and image-space reconstruction losses for paired LR-to-HR image super-resolution. Following the development of the shared supervised super-resolution adapter, FoA-SR generates a shared stochastic candidate pool for each input image and ranks the same candidates using profile-specific Faithful and Aesthetic rewards to mine winner--loser pairs. These pairs are used to fine-tune separate LoRA adapters while keeping the base model frozen. Experiments on RealSR and DIV2K show that FoA-SR can steer the same SR adapter towards distinct restoration objectives: a Faithful adapter improves reference-consistent metrics while an Aesthetic adapter boosts metrics that measure perceptual quality without reference. Our candidate-pool analysis shows that Faithful and Aesthetic rewards frequently select different winners, and a Hybrid-LoRA ablation shows that collapsing both profiles into one reward yields an implicit compromise rather than explicit profile control.
\end{abstract}

\begin{figure}[t]
\centering
\includegraphics[width=0.95\linewidth]{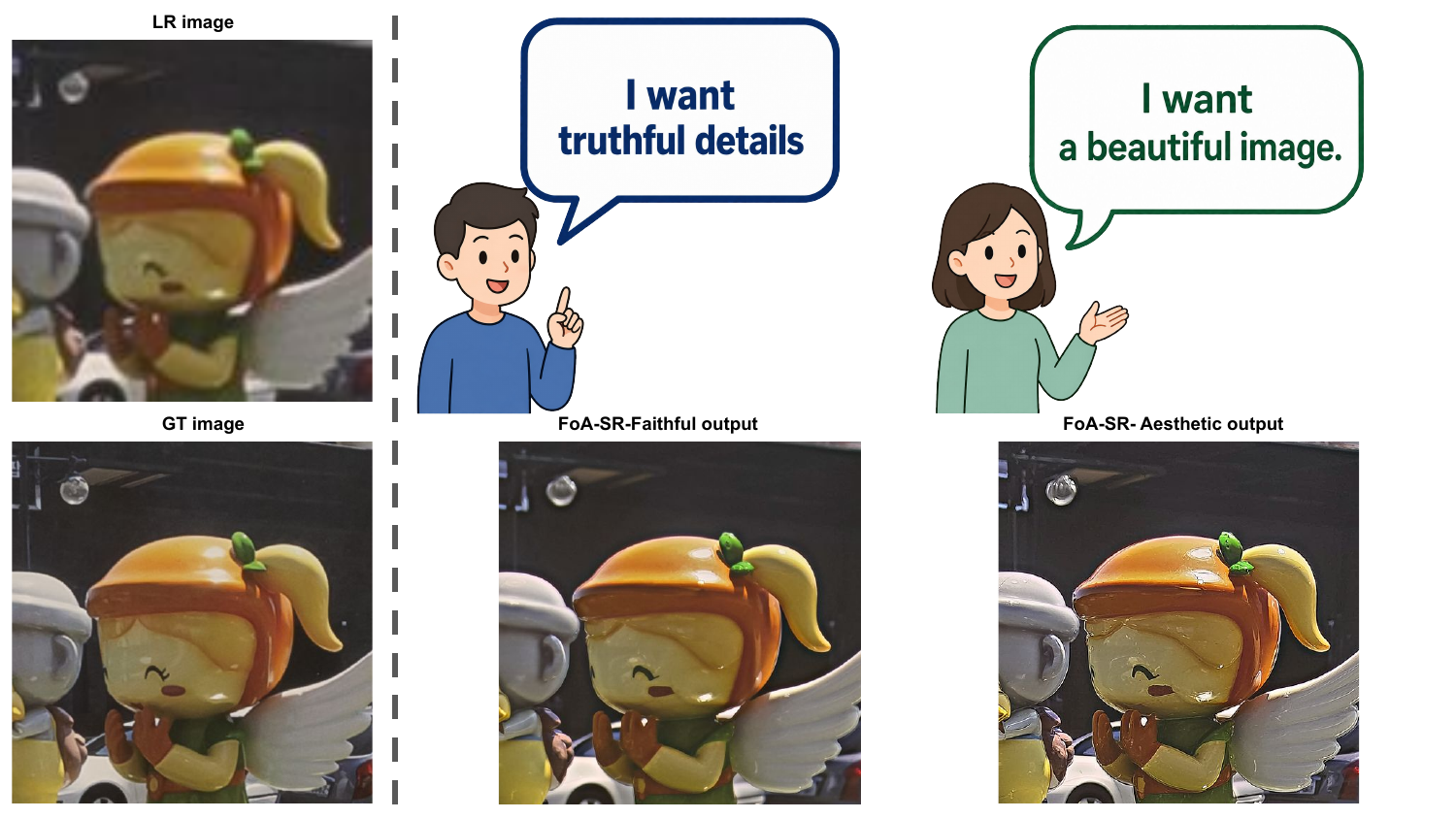}
\caption{Qualitative illustration of profile-dependent restoration. Given the same LR input, FoA-SR-Faithful emphasizes reference consistency and reduced hallucination, while FoA-SR-Aesthetic emphasizes visually pleasing textures and natural-looking details.}
\label{fig:qualitative}
\end{figure}

\section{Introduction}

Real-world image super-resolution (Real-ISR) \citep{Dong2014LearningAD, zhang2018rcan, Liang_2021_ICCV, zhang2022efficientlongrangeattentionnetwork, chen2023activating, li2023efficient, zhang2024transcending, guo2024mambair} aims to recover a high-resolution image from a degraded low-resolution observation. The problem is inherently ill-posed: many high-resolution images can explain the same low-resolution input, especially under unknown blur, noise, compression, and downsampling~\citep{cai2019realsr, zhang2021bsrgan, wang2021realesrgan, blau2018perception}. Recent generative image models have improved the visual realism of restoration systems by synthesizing plausible high-frequency details~\citep{wang2023stablesr, yang2023pasd, lin2024diffbir, wu2024seesr, sun2024ccsr, wu2024osediff, dong2025tsdsr}. 
However, this generative capability also exposes a key ambiguity: for the same low-resolution input, there may be multiple plausible restorations, and the preferred one depends on the intended restoration behavior rather than on a single universal notion of quality~\citep{blau2018perception, wu2025dp2osr}.

Most SR pipelines still optimize and evaluate models as if one output should be best for all criteria. 
This assumption is limiting because perceptual quality and distortion fidelity can conflict~\citep{blau2018perception}. 
A restoration that is visually rich and natural-looking may introduce hallucinated details or alter fine structures, while a conservative restoration may better preserve the reference structure but appear less sharp or less aesthetically pleasing~\citep{wu2025dp2osr}. 
We therefore distinguish between two complementary restoration profiles. 
A \emph{Faithful} restoration prioritizes reference consistency, structure preservation, and reduced hallucination. 
An \emph{Aesthetic} restoration prioritizes visually pleasing details and no-reference perceptual quality. 
These profiles are not merely two names for different metrics; they can induce different rankings over the same set of candidate SR outputs.

This distinction is especially important for large generative backbones. 
FLUX.2-dev is a 32B open-weight rectified-flow transformer for image generation and editing, rather than a paired LR-to-HR super-resolution model~\citep{blackforestlabs2025flux2, blackforestlabs2026flux2dev}. 
Adapting such a model to Real-ISR requires more than applying a generic image-to-image script: the low-resolution image must be represented as a conditioning signal, the high-resolution image must define the restoration target, and the sampling path must remain tied to the LR input rather than starting from unconstrained pure noise. 
To the best of our knowledge, FoA-SR is among the first few works to explore FLUX.2-based rectified-flow backbones for real-world image super-resolution~\citep{strsr}, and the first to study Faithful-versus-Aesthetic preference optimization on top of a full 32B FLUX.2-dev paired LR-to-HR SR adapter. 
In this work, we build a supervised FLUX.2-dev-based SR adapter, which we call \textbf{Flux2SR}, by conditioning the transformer on packed LR latents and training LoRA parameters with a flow-matching objective toward HR latents, together with SR-specific image-space reconstruction losses. 
Flux2SR provides the shared SR baseline from which different restoration preferences can be studied.

However, supervised SR training still produces a single compromise model. Recent preference-optimization methods show that stochastic generative outputs can be ranked and converted into preference pairs for post-training alignment~\citep{rafailov2023dpo, wallace2024diffdpo, wu2025dp2osr}. Yet aligning to a single perceptual or hybrid reward can obscure conflicting restoration intents: Faithful restoration favors identity, text, and structural preservation, while Aesthetic restoration favors visually pleasing enhancement. Instead of collapsing these objectives into one scalar reward, FoA-SR makes the restoration profile explicit.

We propose \textbf{FoA-SR}, a profile-aware preference optimization framework for Real-ISR. Starting from the shared supervised FLUX.2 SR adapter (Flux2SR), FoA-SR samples multiple stochastic candidates for each low-resolution input. The same candidate pool is then scored using profile-specific automatic rewards. The Faithful reward emphasizes full-reference consistency metrics, while the Aesthetic reward emphasizes no-reference perceptual quality metrics. For each profile, FoA-SR constructs winner--loser preference pairs and fine-tunes a profile-specific LoRA adapter. Because both profiles start from the same SR baseline and the same candidate pool, the resulting behavior is driven by the preference profile rather than by different data or model initialization.

Experiments on RealSR and DIV2K show that FoA-SR steers the same Flux2SR baseline toward distinct restoration behaviors. 
The Faithful adapter improves reference-oriented reconstruction, whereas the Aesthetic adapter improves no-reference perceptual quality while sacrificing strict fidelity. 
This specialization supports our central claim that Real-ISR preference is profile-dependent and should be exposed rather than collapsed into a single objective. On a 500-image mining set, Faithful and Aesthetic rewards select different winners for 78.4\% of inputs, showing that the two profiles induce genuinely different preferences over the same stochastic candidate pool.

Our contributions are summarized as follows:
\begin{itemize}
    \item We introduce \textbf{Flux2SR}, a full FLUX.2-dev paired real-world image super-resolution adapter using LR latent conditioning, flow-matching LoRA training, and SR-specific reconstruction losses.
    \item We formulate Real-ISR as a \emph{profile-aware preference optimization} problem, where Faithful and Aesthetic restoration represent distinct user intents rather than a single universal objective.
    \item We propose \textbf{FoA-SR}, which mines profile-specific winner--loser pairs from a shared stochastic candidate pool using automatic Faithful and Aesthetic rewards, without human preference labels.
    \item We show profile specialization on RealSR and DIV2K, and provide profile-disagreement and Hybrid-LoRA analyses demonstrating that collapsing both profiles into one reward hides useful controllability.
\end{itemize}

\section{Related Work}

\paragraph{Real-world image super-resolution.}
Single-image super-resolution has evolved from distortion-oriented reconstruction models to perceptual and real-world restoration systems. Early CNN- and Transformer-based SR methods optimize reconstruction accuracy under simplified degradations~\citep{Dong2014LearningAD, zhang2018rcan, Liang_2021_ICCV, zhang2022efficientlongrangeattentionnetwork, chen2023activating, li2023efficient, zhang2024transcending, guo2024mambair}. However, real-world degradations are unknown and often include blur, noise, compression, and sensor artifacts. Blind and real-world SR methods such as BSRGAN and Real-ESRGAN address this by designing more practical degradation pipelines and perceptual training objectives~\citep{zhang2021bsrgan, wang2021realesrgan}. These methods improve realism but still generally produce a single restoration behavior for each model.

\paragraph{Generative priors for real-world SR.}
Recent work has introduced diffusion and large text-to-image priors into real-world SR, enabling models to synthesize more realistic high-frequency details. StableSR, PASD, DiffBIR, SeeSR, and CCSR exploit generative priors for blind or semantics-aware restoration~\citep{wang2023stablesr, yang2023pasd, lin2024diffbir, wu2024seesr, sun2024ccsr}. OSEDiff and TSD-SR further improve efficiency by distilling diffusion-based restoration into fewer sampling steps~\citep{wu2024osediff, dong2025tsdsr}. These methods demonstrate the power of generative backbones, but they typically aim to produce one preferred output per model. In contrast, FoA-SR treats the desired restoration behavior as profile-dependent and explicitly separates Faithful and Aesthetic outputs.

\paragraph{Preference optimization for generative models.}
Preference optimization has become an effective alternative to reward-model-based alignment. Direct Preference Optimization (DPO) directly optimizes a policy from preference pairs without training a separate reward model~\citep{rafailov2023dpo}, and Diff-DPO extends this idea to diffusion models~\citep{wallace2024diffdpo}. Closest to our work, DP$^2$O-SR exploits stochastic SR outputs, ranks them using IQA-based rewards, constructs preference pairs, and performs post-training preference optimization for Real-ISR~\citep{wu2025dp2osr}. However, DP$^2$O-SR combines full-reference and no-reference signals into a single perceptual or hybrid reward. FoA-SR takes a different view: reference consistency and no-reference perceptual quality correspond to different restoration intents. Rather than merging them into one reward, FoA-SR preserves them as Faithful and Aesthetic profiles and learns separate adapters for each.

\paragraph{Profile-dependent restoration quality.}
The conflict between distortion fidelity and perceptual quality is closely related to the perception--distortion trade-off~\citep{blau2018perception}. A model that maximizes perceptual realism may not preserve the reference exactly, while a model optimized for distortion metrics may produce conservative or less visually appealing outputs. FoA-SR operationalizes this trade-off as a profile-aware preference optimization problem. Instead of treating the trade-off as a single fixed balance, it exposes the restoration intent as an explicit choice between Faithful and Aesthetic adapters.

\section{Method}
\label{sec:method}
\begin{figure}[t]
\centering
\includegraphics[width=\linewidth]{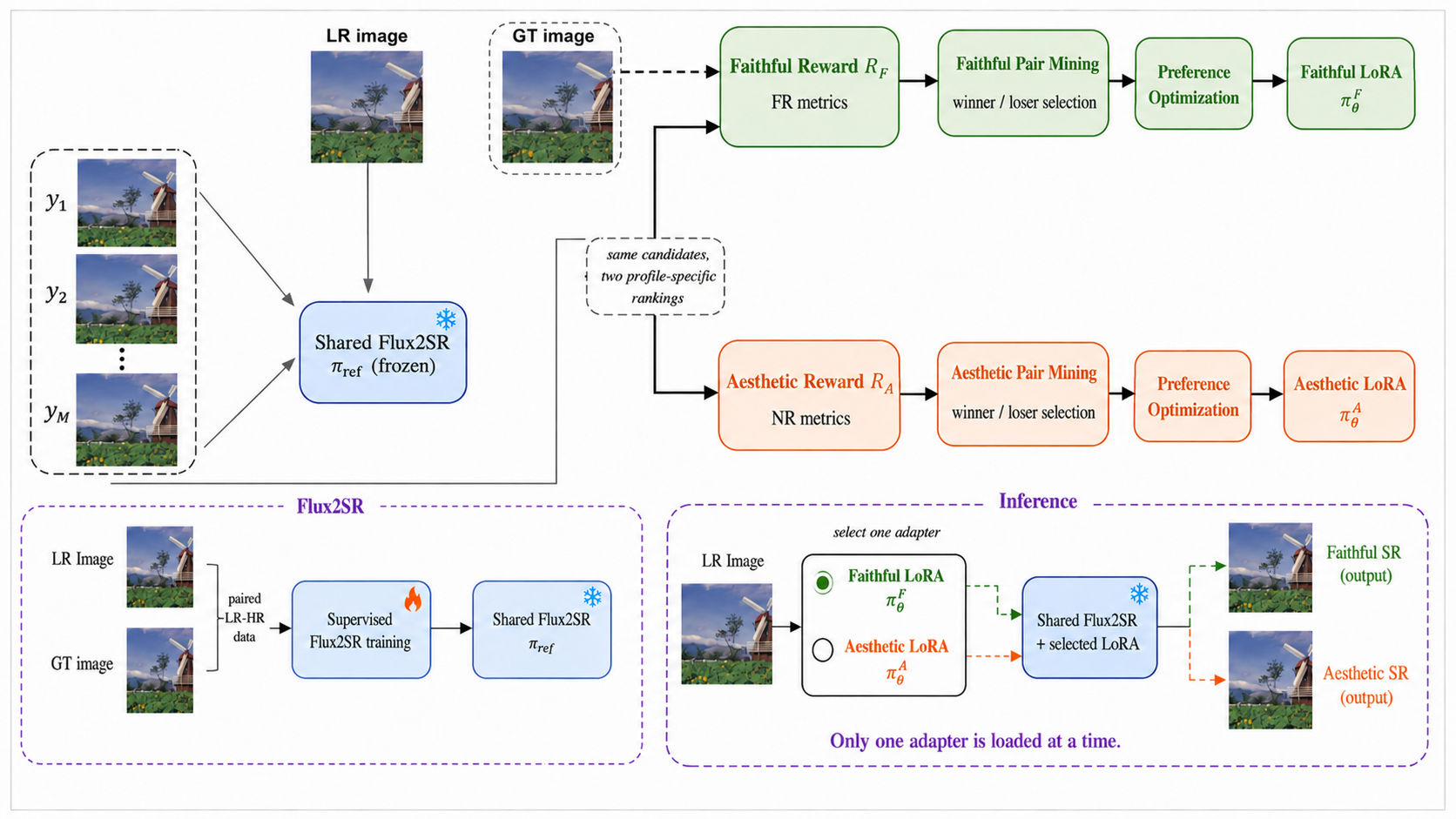}
\caption{
Overview of the proposed FoA-SR framework. Stage 1 adapts FLUX.2-dev into a shared supervised SR baseline, Flux2SR, using LR-HR paired supervision. Stage 2 uses the same Flux2SR model to generate a shared stochastic candidate pool for each LR input. The same candidates are ranked separately by Faithful and Aesthetic rewards to mine profile-specific winner--loser pairs. FoA-SR then fine-tunes separate LoRA adapters for the two profiles. At inference time, the user selects either the Faithful or Aesthetic adapter, exposing the restoration trade-off as a controllable choice rather than collapsing it into a single output.
}
\label{fig:framework}
\end{figure}

FoA-SR consists of two stages. First, we adapt FLUX.2-dev into a paired real-world super-resolution model, which we call \textbf{Flux2SR}. Second, we use Flux2SR as a shared stochastic reference model to construct profile-specific preference pairs and fine-tune separate Faithful and Aesthetic adapters. The key difference from single-reward preference alignment is that FoA-SR does not collapse restoration quality into one universal objective; instead, it preserves Faithful and Aesthetic restoration as separate preference profiles. Figure~\ref{fig:framework} illustrates the overall FoA-SR pipeline.

\subsection{Flux2SR: Adapting FLUX.2-dev for paired super-resolution}

FLUX.2-dev is a large rectified-flow image generation backbone, but it is not directly a paired LR-to-HR super-resolution model~\citep{blackforestlabs2025flux2, blackforestlabs2026flux2dev, huggingface2025flux2diffusers}.  We therefore build \textbf{Flux2SR}, a supervised SR adapter that conditions the FLUX.2 transformer on the low-resolution image while using the high-resolution image as the restoration target. Given an LR-HR pair $(x, y^\star)$, we encode both images into the FLUX.2 latent space. The LR latent is packed as a conditioning image sequence, while the HR latent defines the flow-matching target.

We fine-tune only LoRA parameters in the FLUX.2 transformer~\citep{hu2021lora} and keep the VAE, text encoder, and base transformer weights frozen. The supervised SR objective combines latent flow matching with image-space reconstruction losses:
\begin{equation}
    \mathcal{L}_{\mathrm{SR}}
    =
    \mathcal{L}_{\mathrm{FM}}
    +
    \lambda_{\mathrm{pix}}\mathcal{L}_{\mathrm{charb}}
    +
    \lambda_{\mathrm{lpips}}\mathcal{L}_{\mathrm{LPIPS}}.
\end{equation}
Here, $\mathcal{L}_{\mathrm{FM}}$ trains the LR-conditioned flow trajectory toward the HR latent, while the Charbonnier and LPIPS losses encourage image-space reconstruction quality. Flux2SR serves as the shared supervised baseline and frozen reference model for subsequent preference optimization.

\subsection{Profile-Specific Reward Design and Preference Mining}

Given an LR input $x$, the shared Flux2SR baseline generates a stochastic candidate set
\begin{equation}
    \mathcal{Y}(x)=\{y_1,\ldots,y_M\},
\end{equation}
by varying the random seed. Following recent preference-based Real-ISR work, we exploit these stochastic SR candidates as a source of preference supervision~\citep{wu2025dp2osr}. FoA-SR mines preference pairs from this same candidate pool under two restoration profiles: \emph{Faithful}, which favors reference consistency and reduced hallucination, and \emph{Aesthetic}, which favors visually pleasing no-reference perceptual quality.

To score candidates, we first direction-align and normalize each image-quality metric within the candidate set. For a metric $m$, let $s_i^m$ denote the direction-aligned score of candidate $y_i$, where higher values always indicate better quality. We compute
\begin{equation}
    \bar{s}_i^m
    =
    \frac{s_i^m - \min_j s_j^m}
    {\max_j s_j^m - \min_j s_j^m + \epsilon},
\end{equation}
where $\epsilon$ is a small constant for numerical stability. For lower-is-better metrics such as LPIPS, DISTS, and NIQE, we reverse the metric direction before normalization.

The Faithful reward combines full-reference metrics, including PSNR, SSIM~\citep{wang2004ssim}, LPIPS~\citep{zhang2018lpips}, and DISTS~\citep{ding2020dists}:
\begin{equation}
\begin{aligned}
    R_{\mathrm{F}}(y_i)
    =
    &0.30\,\bar{s}_i^{\mathrm{PSNR}}
    +0.30\,\bar{s}_i^{\mathrm{SSIM}} \\
    &+0.20\,\bar{s}_i^{\mathrm{LPIPS}}
    +0.20\,\bar{s}_i^{\mathrm{DISTS}} .
\end{aligned}
\end{equation}
The Aesthetic reward combines no-reference metrics, including CLIPIQA~\citep{wang2023clipiqa}, MUSIQ~\citep{ke2021musiq}, MANIQA~\citep{yang2022maniqa}, and NIQE~\citep{mittal2013niqe}:
\begin{equation}
\begin{aligned}
    R_{\mathrm{A}}(y_i)
    =
    &0.25\,\bar{s}_i^{\mathrm{CLIPIQA}}
    +0.25\,\bar{s}_i^{\mathrm{MUSIQ}} \\
    &+0.25\,\bar{s}_i^{\mathrm{MANIQA}}
    +0.25\,\bar{s}_i^{\mathrm{NIQE}} .
\end{aligned}
\end{equation}

For each profile $p \in \{\mathrm{F},\mathrm{A}\}$, we select the highest-scoring candidate as the winner and the lowest-scoring candidate as the loser:
\begin{equation}
    y_w^p=\arg\max_{y_i\in\mathcal{Y}(x)} R_p(y_i),
    \qquad
    y_l^p=\arg\min_{y_i\in\mathcal{Y}(x)} R_p(y_i).
\end{equation}
The reward gap
\begin{equation}
    \Delta_p = R_p(y_w^p)-R_p(y_l^p)
\end{equation}
is used as a confidence score, and low-gap pairs are filtered out during preference-data construction.

This procedure yields two profile-specific preference datasets, one for Faithful restoration and one for Aesthetic restoration. Importantly, both datasets are mined from the same Flux2SR candidate pool. Therefore, when the two profiles select different winners for the same input, the disagreement reflects a restoration-profile preference rather than a difference in data, sampling source, or model initialization. Unlike a hybrid reward that averages reference and no-reference criteria into a single scalar objective, FoA-SR preserves the two rankings and uses them to learn specialized restoration adapters.
\paragraph{Mining versus evaluation.}
The within-pool normalization in Eq.~(3) is used only for candidate ranking and preference-pair mining within the same LR input. 
It is not used when comparing final models. 
All final model comparisons in Table~\ref{tab:main_all_realsr_div2k} are reported using raw, non-normalized evaluation metrics on the test sets. 
In addition to the metrics used to construct the Faithful and Aesthetic rewards, we report non-optimized perceptual metrics, including TOPIQ-FR, AFINE-FR, CLIPIQA+, TOPIQ-NR, AFINE-NR, NIMA, and TOPIQ-IAA. 
These metrics provide an additional check that the learned profiles generalize beyond the direct reward metrics used for preference mining.
\subsection{Preference Optimization for Faithful and Aesthetic Adapters}

For each profile $p$, we initialize a trainable LoRA adapter $\pi_{\theta_p}$ from Flux2SR and use the original Flux2SR adapter as the frozen reference policy $\pi_{\mathrm{ref}}$. We then optimize $\pi_{\theta_p}$ using the mined preference pairs $(x,y_w^p,y_l^p)$.

We optimize the profile-specific adapters with a DPO-style objective~\citep{rafailov2023dpo}, following its adaptation to diffusion and flow-based generative models~\citep{wallace2024diffdpo, wu2025dp2osr}. Because FLUX.2 is a flow-matching model, we use the flow-matching reconstruction loss as a negative log-likelihood surrogate. Let $\ell_{\theta_p}(y\mid x)$ denote the flow-matching loss of policy $\pi_{\theta_p}$ when reconstructing candidate $y$ under LR condition $x$, and let $\ell_{\mathrm{ref}}(y\mid x)$ denote the corresponding loss under the frozen reference model. We define the preference margins
\begin{equation}
    A_{\theta_p}
    =
    \ell_{\theta_p}(y_l^p\mid x)
    -
    \ell_{\theta_p}(y_w^p\mid x),
    \qquad
    A_{\mathrm{ref}}
    =
    \ell_{\mathrm{ref}}(y_l^p\mid x)
    -
    \ell_{\mathrm{ref}}(y_w^p\mid x).
\end{equation}
The profile-specific preference loss is
\begin{equation}
    \mathcal{L}_{\mathrm{pref}}^p
    =
    -
    \mathbb{E}_{(x,y_w^p,y_l^p)}
    \left[
    \log \sigma
    \left(
    \beta \left(A_{\theta_p}-A_{\mathrm{ref}}\right)
    \right)
    \right],
\end{equation}
where $\beta$ controls the strength of preference optimization. This objective encourages the profile-specific adapter to assign lower flow-matching loss to the winner than to the loser, relative to the frozen Flux2SR reference.

After optimization, FoA-SR yields two adapters: \textbf{FoA-SR-Faithful}, which is aligned with reference-consistent restoration, and \textbf{FoA-SR-Aesthetic}, which is aligned with no-reference perceptual quality. At inference time, the user selects the adapter corresponding to the desired restoration profile.
\section{Experiments}

\begin{table}[t]
\centering
\scriptsize
\caption{
Performance comparison of different Real-ISR methods on RealSR and DIV2K-Val. 
Metric types are categorized into training-aligned FR metrics 
(\textcolor{blue}{blue}), training-aligned NR metrics 
(\textcolor{green!60!black}{green}), non-optimized FR perceptual metrics 
(\textcolor{purple}{purple}), non-optimized NR metrics 
(\textcolor{orange!90!black}{orange}). 
\textcolor{red}{Red bold} values indicate better performance
between the baseline Flux2SR and our variants FoA-SR-Faithful and FoA-SR-Aesthetic. 
Arrows indicate whether higher ($\uparrow$) or lower ($\downarrow$) values are better.
}
\label{tab:main_all_realsr_div2k}
\setlength{\tabcolsep}{3.6pt}
\renewcommand{\arraystretch}{1.05}
\resizebox{\linewidth}{!}{
\begin{tabular}{llcccccccc}
\toprule
Dataset & Metric 
& OSEDiff 
& C-SD2 
& DP$^2$O-SD2 
& C-FLUX 
& DP$^2$O-FLUX 
& Flux2SR 
& FoA-Faithful 
& FoA-Aesthetic \\
\midrule

\rowcolor{trainfr}
\multirow{16}{*}{\textit{RealSR}} & PSNR $\uparrow$ 
& 25.1511 & 22.8531 & 21.9559 & 24.5010 & 24.5504 & 22.7195 & \redbest{24.8633} & 20.9118 \\

\rowcolor{trainfr}
& SSIM $\uparrow$ 
& 0.7341 & 0.6026 & 0.6212 & 0.6807 & 0.6785 & 0.6733 & \redbest{0.7327} & 0.5695 \\

\rowcolor{trainfr}
& LPIPS $\downarrow$ 
& 0.2920 & 0.4247 & 0.3810 & 0.3328 & 0.3253 & 0.2793 & \redbest{0.2502} & 0.4038 \\

\rowcolor{trainfr}
& DISTS $\downarrow$ 
& 0.2128 & 0.2694 & 0.2589 & 0.2225 & 0.2276 & 0.2191 & \redbest{0.2046} & 0.2743 \\

\rowcolor{trainnr}
& CLIPIQA $\uparrow$ 
& 0.6688 & 0.7068 & 0.7519 & 0.6312 & 0.7127 & 0.6154 & 0.5535 & \redbest{0.7170} \\

\rowcolor{trainnr}
& NIQE $\downarrow$ 
& 5.6359 & 6.7806 & 5.8413 & 5.0952 & 4.6720 & \redbest{4.9360} & 5.1734 & 5.0522 \\

\rowcolor{trainnr}
& MUSIQ $\uparrow$ 
& 69.0897 & 68.7853 & 72.3349 & 69.7476 & 72.5801 & 69.5302 & 66.6408 & \redbest{73.0874} \\

\rowcolor{trainnr}
& MANIQA $\uparrow$ 
& 0.6335 & 0.6538 & 0.6996 & 0.6673 & 0.6892 & 0.6667 & 0.6252 & \redbest{0.7270} \\

\rowcolor{nonoptfr}
& TOPIQ-FR $\uparrow$ 
& 0.5059 & 0.4469 & 0.4436 & 0.4869 & 0.4932 & 0.4769 & \redbest{0.5146} & 0.4118 \\

\rowcolor{nonoptfr}
& AFINE-FR $\downarrow$ 
& -0.7171 & -0.6284 & -0.3500 & -0.6041 & -0.6467 & -0.7410 & \redbest{-0.9138} & -0.4605 \\

\rowcolor{nonoptnr}
& CLIPIQA+ $\uparrow$ 
& 0.6963 & 0.6860 & 0.7589 & 0.6772 & 0.7408 & 0.6792 & 0.6380 & \redbest{0.7579} \\

\rowcolor{nonoptnr}
& TOPIQ-NR $\uparrow$ 
& 0.6249 & 0.6553 & 0.7365 & 0.6528 & 0.7345 & 0.6258 & 0.5847 & \redbest{0.7408} \\

\rowcolor{nonoptnr}
& AFINE-NR $\downarrow$ 
& -1.0488 & -0.9824 & -1.1185 & -1.0823 & -1.0659 & -1.0478 & -0.9974 & \redbest{-1.0772} \\

\rowcolor{nonoptnr}
& NIMA $\uparrow$ 
& 4.8939 & 4.9534 & 5.1149 & 4.9310 & 5.0096 & 4.8029 & 4.6323 & \redbest{5.1457} \\

\rowcolor{nonoptnr}
\multirow{-16}{*}{\textit{RealSR}}
& TOPIQ-IAA $\uparrow$ 
& 4.7536 & 4.8699 & 4.9881 & 4.7115 & 4.9006 & 4.6670 & 4.5808 & \redbest{5.0062} \\

\midrule

\rowcolor{trainfr}
\multirow{16}{*}{\textit{DIV2K-Val}} & PSNR $\uparrow$
& 20.2656 & 19.3126 & 18.2443 & 19.2213 & 19.3625 & 18.6927 & \redbest{19.6302} & 17.5904 \\

\rowcolor{trainfr}
& SSIM $\uparrow$ 
& 0.4851 & 0.4166 & 0.3879 & 0.4350 & 0.4391 & 0.4386 & \redbest{0.4758} & 0.3762 \\

\rowcolor{trainfr}
& LPIPS $\downarrow$ 
& 0.3852 & 0.4472 & 0.4462 & 0.4174 & 0.3955 & \redbest{0.4024} & 0.4061 & 0.4216 \\

\rowcolor{trainfr}
& DISTS $\downarrow$ 
& 0.2392 & 0.2669 & 0.2605 & 0.2324 & 0.2306 & \redbest{0.2279} & 0.2306 & 0.2506 \\

\rowcolor{trainnr}
& CLIPIQA $\uparrow$ 
& 0.6572 & 0.7340 & 0.7812 & 0.6792 & 0.7593 & 0.5673 & 0.5074 & \redbest{0.7367} \\

\rowcolor{trainnr}
& NIQE $\downarrow$ 
& 4.1857 & 5.4295 & 4.7417 & 4.1298 & 4.1416 & 4.5140 & 4.8010 & \redbest{4.2739} \\

\rowcolor{trainnr}
& MUSIQ $\uparrow$ 
& 67.4412 & 68.7474 & 73.4377 & 69.1386 & 73.2998 & 68.1995 & 64.9050 & \redbest{73.9127} \\

\rowcolor{trainnr}
& MANIQA $\uparrow$ 
& 0.6031 & 0.6537 & 0.7122 & 0.6611 & 0.6998 & 0.6589 & 0.6098 & \redbest{0.7385} \\

\rowcolor{nonoptfr}
& TOPIQ-FR $\uparrow$ 
& 0.3685 & 0.3466 & 0.3539 & 0.3609 & 0.3709 & \redbest{0.3575} & 0.3556 & 0.3356 \\

\rowcolor{nonoptfr}
& AFINE-FR $\downarrow$ 
& -0.4953 & -0.3917 & -0.4740 & -0.5211 & -0.5845 & \redbest{-0.6603} & -0.5858 & -0.5924 \\

\rowcolor{nonoptnr}
& CLIPIQA+ $\uparrow$ 
& 0.6852 & 0.6972 & 0.7720 & 0.7125 & 0.7783 & 0.6855 & 0.6518 & \redbest{0.7867} \\

\rowcolor{nonoptnr}
& TOPIQ-NR $\uparrow$ 
& 0.6075 & 0.6546 & 0.7346 & 0.6756 & 0.7545 & 0.5864 & 0.5446 & \redbest{0.7600} \\

\rowcolor{nonoptnr}
& AFINE-NR $\downarrow$ 
& -0.8597 & -0.9168 & -1.0908 & -1.0520 & -1.1076 & -0.9522 & -0.8939 & \redbest{-1.0534} \\

\rowcolor{nonoptnr}
& NIMA $\uparrow$ 
& 4.9651 & 5.2102 & 5.4259 & 5.1806 & 5.3697 & 4.8478 & 4.6236 & \redbest{5.3094} \\

\rowcolor{nonoptnr}
\multirow{-16}{*}{\textit{DIV2K-Val}}
& TOPIQ-IAA $\uparrow$ 
& 4.8434 & 5.0965 & 5.2542 & 4.9838 & 5.2233 & 4.6482 & 4.5429 & \redbest{5.1901} \\

\bottomrule
\end{tabular}
}
\end{table}

\subsection{Experimental Settings}

\paragraph{Training and testing datasets.}
Following the training setup commonly used in recent Real-ISR work, we train Flux2SR using the LSDIR dataset~\citep{lsdir} and the first 10K face images from FFHQ~\citep{karras2019ffhq}. 
We synthesize paired LR-HR training data using the Real-ESRGAN degradation pipeline~\citep{wang2021realesrgan}. 
This produces realistic low-quality inputs with diverse blur, noise, compression, and downsampling artifacts, paired with their corresponding high-quality targets. 
Each training sample is further cached as FLUX.2 VAE latents for efficient flow-matching training.

For profile-specific preference optimization, we randomly sample 500 LR-HR pairs from the training data as a pilot mining set. 
Starting from the shared Flux2SR baseline, we generate $M=4$ stochastic SR candidates for each LR input by varying the random seed. 
The same candidate pool is scored by the Faithful and Aesthetic rewards, producing two profile-specific winner--loser preference datasets used to train the FoA-SR-Faithful and FoA-SR-Aesthetic adapters. 
We analyze the effect of different candidate pool sizes in Section~\ref{sec:analysis}, with detailed statistics in Appendix~\ref{app:mining}.

For evaluation, we report results on RealSR~\citep{cai2019realsr} and DIV2K-Val~\citep{agustsson2017ntire}. 
For DIV2K-Val, we follow the common synthetic Real-ISR evaluation protocol by randomly cropping 100 HR images and degrading them using the Real-ESRGAN degradation pipeline. 
RealSR provides real-world LR-HR pairs, while DIV2K-Val provides a controlled synthetic benchmark with known HR references. 
All methods are evaluated on the same LR inputs and HR references using the same metric implementation.

\paragraph{Compared Methods.}
We evaluate FoA-SR against Real-ISR baselines and our Flux2SR-based variants under the same evaluation protocol. 
External baselines include OSEDiff~\citep{wu2024osediff}, C-SD2, DP$^2$O-SD2, C-FLUX, and DP$^2$O-FLUX~\citep{wu2025dp2osr}. 
Our internal variants include the shared supervised baseline \textbf{Flux2SR}, \textbf{FoA-SR-Faithful}, and \textbf{FoA-SR-Aesthetic}. 
We use \textbf{FoA-SR-Hybrid} only for ablation analysis with the averaged reward $R_H=0.5R_F+0.5R_A$.

\paragraph{Evaluation metrics.}
First, we report training-aligned full-reference metrics used by the Faithful reward: PSNR, SSIM~\citep{wang2004ssim}, LPIPS~\citep{zhang2018lpips}, and DISTS~\citep{ding2020dists}. 
Second, we report training-aligned no-reference metrics used by the Aesthetic reward: CLIPIQA~\citep{wang2023clipiqa}, NIQE~\citep{mittal2013niqe}, MUSIQ~\citep{ke2021musiq}, and MANIQA~\citep{yang2022maniqa}. 
Third, we report non-optimized full-reference perceptual metrics that are not used for preference mining, including TOPIQ-FR and AFINE-FR. 
Finally, we report non-optimized no-reference and aesthetic metrics, including CLIPIQA+, TOPIQ-NR, AFINE-NR, NIMA, and TOPIQ-IAA. All final model comparisons are reported using raw, non-normalized metric values.

\paragraph{Implementation details.}
All Flux2SR-based models are built on FLUX.2-dev. For Flux2SR, we freeze the VAE, text encoder, and base transformer weights, and fine-tune only LoRA parameters in the transformer. Unless otherwise stated, Flux2SR is trained with LoRA rank 64, AdamW optimization, learning rate $1\times10^{-4}$, weight decay $1\times10^{-4}$, and a constant-with-warmup learning-rate schedule.

For profile-specific preference optimization, both FoA-SR-Faithful and FoA-SR-Aesthetic are initialized from the same Flux2SR checkpoint. We fine-tune only LoRA parameters and keep the FLUX.2 backbone frozen. Preference training uses mined winner--loser latent pairs, learning rate $2\times10^{-5}$, LoRA rank 64, and DPO preference strength $\beta=1000$. The 500-image mining set is used only for preference-pair construction and analysis of the candidate pool size; it is not used as a separate evaluation benchmark. For fair evaluation, all methods are evaluated on the same LR inputs and HR references using the same metric implementation. Additional training hyperparameters, compute resources, and evaluation protocol details are provided in Appendix~\ref{app:implementation}.
\subsection{Main Results and Profile Specialization}

\paragraph{Quantitative comparison.}
Table~\ref{tab:main_all_realsr_div2k} reports the main quantitative comparison on RealSR and DIV2K-Val under the same evaluation protocol. 
External baselines provide reference points for current Real-ISR performance, while the red-bold internal comparison isolates the effect of profile-specific preference optimization from the same Flux2SR starting point.

Across both datasets, FoA-SR-Faithful improves distortion-oriented reference metrics, especially PSNR and SSIM. 
On RealSR, it also improves perceptual reference metrics such as LPIPS, DISTS, TOPIQ-FR, and AFINE-FR. 
In contrast, FoA-SR-Aesthetic improves learned no-reference and aesthetic metrics, including CLIPIQA, MUSIQ, MANIQA, CLIPIQA+, TOPIQ-NR, NIMA, and TOPIQ-IAA, while sacrificing strict reference fidelity. 
These results show that FoA-SR does not optimize a single universal quality score; instead, it steers the same Flux2SR baseline toward distinct restoration profiles.

Compared with single-output Real-ISR baselines such as DP$^2$O-SR and OSEDiff, FoA-SR provides a complementary capability: it exposes restoration intent as an explicit choice between Faithful and Aesthetic adapters rather than committing to one fixed balance between fidelity and perceptual enhancement.

\begin{figure}[t]
\centering
\includegraphics[width=\linewidth]{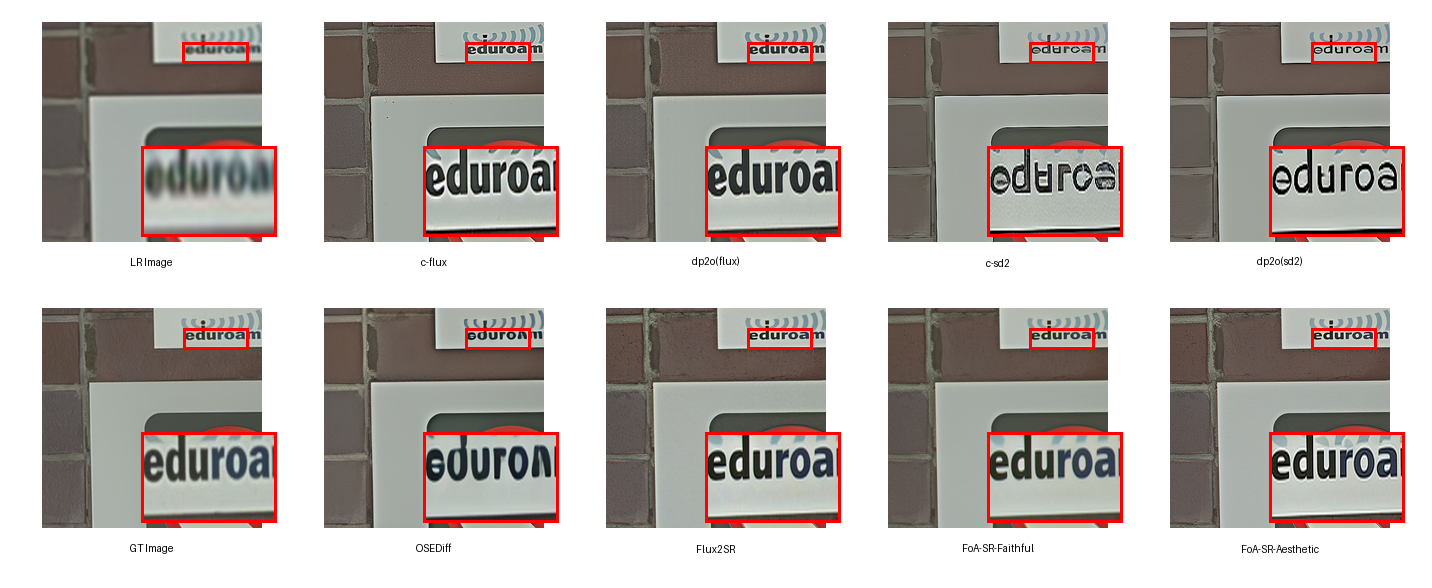}
\caption{Qualitative comparison on a text-containing RealSR example. FoA-SR-Faithful emphasizes reference-consistent restoration, while FoA-SR-Aesthetic emphasizes sharper and more visually enhanced details.}
\label{fig:qualitative_eduroam}
\end{figure}

\paragraph{Qualitative comparison.}
Figure~\ref{fig:qualitative_eduroam} shows a qualitative comparison on a text-containing RealSR example. 
FoA-SR-Faithful preserves the text structure more conservatively and avoids strong distortions, while FoA-SR-Aesthetic produces a sharper and more visually enhanced appearance but may introduce stronger local changes.  Additional qualitative comparisons on RealSR and DIV2K-Val are shown in Appendix~\ref{app:qualitative}.

\subsection{Analysis and Ablations}
\label{sec:analysis}

\paragraph{Hybrid reward ablation.}
To test whether Faithful and Aesthetic restoration can be replaced by a single scalar objective, 
we train an additional FoA-SR-Hybrid adapter using the same Flux2SR initialization, candidate pool, and preference optimization setup. 
The only difference is that winner--loser pairs are mined using a hybrid reward:
\[
R_H(y_i) = 0.5R_F(y_i) + 0.5R_A(y_i).
\]
As shown in Table~\ref{tab:hybrid_ablation}, FoA-SR-Hybrid improves several no-reference metrics, including CLIPIQA, NIQE, MUSIQ, and MANIQA, but it does not recover the strong reference-consistency gains achieved by FoA-SR-Faithful. 
This supports our central design choice: averaging Faithful and Aesthetic rewards into a single scalar objective produces an implicit compromise, whereas FoA-SR exposes the two endpoints as selectable restoration profiles. Extended held-out metrics for the Hybrid adapter are reported in Appendix~\ref{app:hybrid_extended}.

\begin{table}[t]
\centering
\scriptsize
\caption{
Hybrid reward ablation on RealSR. FoA-SR-Hybrid is trained with the averaged reward $R_H=0.5R_F+0.5R_A$. 
It improves several no-reference metrics but does not match the reference-consistency gains of FoA-SR-Faithful, supporting profile-specific optimization.
}
\label{tab:hybrid_ablation}
\resizebox{\linewidth}{!}{
\begin{tabular}{lccccccccc}
\toprule
Method 
& PSNR$\uparrow$ 
& SSIM$\uparrow$ 
& LPIPS$\downarrow$ 
& DISTS$\downarrow$ 
& CLIPIQA$\uparrow$ 
& NIQE$\downarrow$ 
& MUSIQ$\uparrow$ 
& MANIQA$\uparrow$ 
& FID$\downarrow$ \\
\midrule
Flux2SR 
& 22.7195 & 0.6733 & 0.2793 & 0.2191 & 0.6154 & 4.9360 & 69.5302 & 0.6667 & 110.0081 \\
FoA-SR-Faithful 
& \textbf{24.8633} & \textbf{0.7327} & \textbf{0.2502} & \textbf{0.2046} & 0.5535 & 5.1734 & 66.6408 & 0.6252 & \textbf{107.1397} \\
FoA-SR-Aesthetic 
& 20.9118 & 0.5695 & 0.4038 & 0.2743 & \textbf{0.7170} & 5.0522 & \textbf{73.0874} & \textbf{0.7270} & 123.7668 \\
FoA-SR-Hybrid 
& 22.6407 & 0.6539 & 0.3252 & 0.2522 & 0.7120 & \textbf{4.5909} & 72.6504 & 0.7026 & 121.1101 \\
\bottomrule
\end{tabular}
}
\end{table}

\paragraph{Do Faithful and Aesthetic profiles disagree?}
We next ask whether the two profiles actually induce different preferences over the same stochastic candidate pool. 
At the training setting used for preference optimization ($M=4$), Faithful and Aesthetic rewards select different winners for 78.4\% of the 500 LR inputs. 
The mean Spearman correlation between the two reward rankings is close to zero ($-0.0312$), indicating that the two profiles do not simply rank candidates in the same order. 
Moreover, each profile assigns a substantial advantage to its own selected winner: the Faithful winner exceeds the Aesthetic winner by 0.3338 under $R_F$, while the Aesthetic winner exceeds the Faithful winner by 0.3411 under $R_A$. 
These results provide direct evidence that Faithful and Aesthetic restoration are genuinely different selection criteria rather than two names for the same automatic preference. Additional disagreement statistics are provided in Appendix~\ref{app:mining}.

\paragraph{Effect of candidate pool size.}
Our main preference-training experiments use $M=4$ candidates per LR input as a compute-efficient setting for FLUX.2-based candidate generation and multi-metric IQA scoring. 
To assess the sensitivity of profile mining to the number of stochastic candidates, we extend the same 500-image mining set up to $M=8$ and recompute profile-mining statistics for $M\in\{2,4,6,8\}$ without retraining the adapters. 
As shown in Table~\ref{tab:candidate_pool_size}, the Faithful and Aesthetic reward gaps are already strong at $M=4$, while larger candidate pools mainly increase winner disagreement. Appendix~\ref{app:mining} reports the corresponding standard deviations and detailed \(M=4\) disagreement statistics.

\begin{table}[t]
\centering
\scriptsize
\caption{
Effect of candidate pool size on profile mining over 500 LR inputs. 
Faithful and Aesthetic rewards frequently select different winners, and reward gaps are already strong at $M=4$, supporting its use as a compute-efficient mining setting.
}
\label{tab:candidate_pool_size}
\begin{tabular}{ccccc}
\toprule
Pool size $M$ 
& Faithful gap $\uparrow$ 
& Aesthetic gap $\uparrow$ 
& Winner disagreement $\uparrow$ 
& Spearman $R_F/R_A$ \\
\midrule
2 & 0.5628 & 0.5700 & 55.2\% & -0.1690 \\
4 & 0.6517 & 0.7097 & 78.4\% & -0.0312 \\
6 & 0.6749 & 0.7216 & 85.4\% & -0.0429 \\
8 & 0.6741 & 0.7300 & 90.8\% & -0.0380 \\
\bottomrule
\end{tabular}
\end{table}

\paragraph{Choice of the shared Flux2SR baseline.}
We use the FM+Charbonnier+LPIPS Flux2SR checkpoint as the shared starting point for all profile-specific adapters. 
This design isolates the effect of the preference profile: FoA-SR-Faithful, FoA-SR-Aesthetic, and FoA-SR-Hybrid all start from the same supervised checkpoint and are trained from candidate pools generated by the same model. 
In preliminary supervised runs, purely reconstruction-oriented baselines improved distortion metrics but tended to produce conservative or overly smooth outputs. 
The FM+Charbonnier+LPIPS Flux2SR baseline provides a more balanced starting point for studying both Faithful and Aesthetic preference optimization. Supervised objective and LoRA target-module ablations supporting this choice are provided in Appendix~\ref{app:extended_results}.
\section{Conclusion and Limitations}

We introduced FoA-SR, a profile-aware preference optimization framework for real-world image super-resolution. 
FoA-SR adapts FLUX.2-dev into a paired SR baseline, Flux2SR, and then mines profile-specific preference pairs from the same stochastic candidate pool to train separate Faithful and Aesthetic LoRA adapters. 
Experiments on RealSR and DIV2K-Val show that FoA-SR steers the same baseline toward distinct restoration behaviors: Faithful improves reference-oriented metrics, while Aesthetic improves learned no-reference and aesthetic metrics. 
Our profile-disagreement analysis and Hybrid-LoRA ablation further show that collapsing both profiles into a single averaged reward can hide useful controllability. 
These results suggest that Real-ISR should not be treated as a single universal restoration objective, but as a profile-dependent restoration problem.

\paragraph{Limitations.}
FoA-SR relies on automatic IQA metrics rather than direct human preference annotations, which may not fully capture subjective visual preference. 
A profile-conditioned human study would provide stronger validation. 
FoA-SR also considers only two discrete profiles, Faithful and Aesthetic, rather than a continuous user-controlled trade-off. 
In addition, preference mining is performed offline and requires generating and scoring multiple candidates per LR input; more efficient online or semi-online mining could reduce this cost. 
Finally, the Aesthetic adapter may introduce local changes or hallucinated details, making the Faithful adapter more appropriate for identity-, text-, document-, or science-sensitive applications.
\bibliographystyle{plainnat}
\bibliography{references}

@article{strsr,
  title={Spectral and Trajectory Regularization for Diffusion Transformer Super-Resolution},
  author={Wang, Jingkai and Tang, Yixin and Gong, Jue and Li, Jiatong and Li, Shu and Liu, Libo and Lan, Jianliang and Liu, Yutong and Zhang, Yulun},
  journal={arXiv preprint arXiv:2603.06275},
  year={2026}
}

@inproceedings{Dong2014LearningAD,
  title={Learning a Deep Convolutional Network for Image Super-Resolution},
  author={Chao Dong and Chen Change Loy and Kaiming He and Xiaoou Tang},
  booktitle={European Conference on Computer Vision},
  year={2014},
  url={https://api.semanticscholar.org/CorpusID:18874645}
}

@inproceedings{zhang2018rcan,
    title={Image Super-Resolution Using Very Deep Residual Channel Attention Networks},
    author={Zhang, Yulun and Li, Kunpeng and Li, Kai and Wang, Lichen and Zhong, Bineng and Fu, Yun},
    booktitle={ECCV},
    year={2018}
}

@InProceedings{Liang_2021_ICCV,
    author    = {Liang, Jingyun and Cao, Jiezhang and Sun, Guolei and Zhang, Kai and Van Gool, Luc and Timofte, Radu},
    title     = {SwinIR: Image Restoration Using Swin Transformer},
    booktitle = {Proceedings of the IEEE/CVF International Conference on Computer Vision (ICCV) Workshops},
    month     = {October},
    year      = {2021},
    pages     = {1833-1844}
}

@misc{zhang2022efficientlongrangeattentionnetwork,
      title={Efficient Long-Range Attention Network for Image Super-resolution}, 
      author={Xindong Zhang and Hui Zeng and Shi Guo and Lei Zhang},
      year={2022},
      eprint={2203.06697},
      archivePrefix={arXiv},
      primaryClass={cs.CV},
      url={https://arxiv.org/abs/2203.06697}, 
}

@inproceedings{chen2023activating,
  title={Activating more pixels in image super-resolution transformer},
  author={Chen, Xiangyu and Wang, Xintao and Zhou, Jiantao and Qiao, Yu and Dong, Chao},
  booktitle={Proceedings of the IEEE/CVF conference on computer vision and pattern recognition},
  pages={22367--22377},
  year={2023}
}

@inproceedings{li2023efficient,
  title={Efficient and explicit modelling of image hierarchies for image restoration},
  author={Li, Yawei and Fan, Yuchen and Xiang, Xiaoyu and Demandolx, Denis and Ranjan, Rakesh and Timofte, Radu and Van Gool, Luc},
  booktitle={Proceedings of the IEEE/CVF conference on computer vision and pattern recognition},
  pages={18278--18289},
  year={2023}
}

@inproceedings{zhang2024transcending,
  title={Transcending the limit of local window: Advanced super-resolution transformer with adaptive token dictionary},
  author={Zhang, Leheng and Li, Yawei and Zhou, Xingyu and Zhao, Xiaorui and Gu, Shuhang},
  booktitle={Proceedings of the IEEE/CVF conference on computer vision and pattern recognition},
  pages={2856--2865},
  year={2024}
}

@inproceedings{guo2024mambair,
  title={Mambair: A simple baseline for image restoration with state-space model},
  author={Guo, Hang and Li, Jinmin and Dai, Tao and Ouyang, Zhihao and Ren, Xudong and Xia, Shu-Tao},
  booktitle={European conference on computer vision},
  pages={222--241},
  year={2024},
  organization={Springer}
}

@inproceedings{cai2019realsr,
  title={Toward Real-World Single Image Super-Resolution: A New Benchmark and a New Model},
  author={Cai, Jianrui and Zeng, Hui and Yong, Hongwei and Cao, Zisheng and Zhang, Lei},
  booktitle={Proceedings of the IEEE/CVF International Conference on Computer Vision},
  pages={3086--3095},
  year={2019}
}

@inproceedings{zhang2021bsrgan,
  title={Designing a Practical Degradation Model for Deep Blind Image Super-Resolution},
  author={Zhang, Kai and Liang, Jingyun and Van Gool, Luc and Timofte, Radu},
  booktitle={Proceedings of the IEEE/CVF International Conference on Computer Vision},
  pages={4791--4800},
  year={2021}
}

@inproceedings{wang2021realesrgan,
  title={Real-ESRGAN: Training Real-World Blind Super-Resolution with Pure Synthetic Data},
  author={Wang, Xintao and Xie, Liangbin and Dong, Chao and Shan, Ying},
  booktitle={Proceedings of the IEEE/CVF International Conference on Computer Vision Workshops},
  pages={1905--1914},
  year={2021}
}

@inproceedings{blau2018perception,
  title={The Perception-Distortion Tradeoff},
  author={Blau, Yochai and Michaeli, Tomer},
  booktitle={Proceedings of the IEEE Conference on Computer Vision and Pattern Recognition},
  pages={6228--6237},
  year={2018}
}

@article{wang2023stablesr,
  title={Exploiting Diffusion Prior for Real-World Image Super-Resolution},
  author={Wang, Jianyi and Yue, Zongsheng and Zhou, Shangchen and Chan, Kelvin C. K. and Loy, Chen Change},
  journal={arXiv preprint arXiv:2305.07015},
  year={2023}
}

@article{yang2023pasd,
  title={Pixel-Aware Stable Diffusion for Realistic Image Super-Resolution and Personalized Stylization},
  author={Yang, Tao and Ren, Peiran and Xie, Xuansong and Zhang, Lei},
  journal={arXiv preprint arXiv:2308.14469},
  year={2023}
}

@inproceedings{lin2024diffbir,
  title={DiffBIR: Toward Blind Image Restoration with Generative Diffusion Prior},
  author={Lin, Xinqi and He, Jingwen and Chen, Ziyan and Lyu, Zhaoyang and Dai, Bo and Yu, Fanghua and Qiao, Yu and Ouyang, Wanli and Dong, Chao},
  booktitle={European Conference on Computer Vision},
  pages={430--448},
  year={2024},
  organization={Springer}
}

@inproceedings{wu2024seesr,
  title={SeeSR: Towards Semantics-Aware Real-World Image Super-Resolution},
  author={Wu, Rongyuan and Yang, Tao and Sun, Lingchen and Zhang, Zhengqiang and Li, Shuai and Zhang, Lei},
  booktitle={Proceedings of the IEEE/CVF Conference on Computer Vision and Pattern Recognition},
  pages={25456--25467},
  year={2024}
}

@article{sun2024ccsr,
  title={Improving the Stability of Diffusion Models for Content Consistent Super-Resolution},
  author={Sun, Lingchen and Wu, Rongyuan and Zhang, Zhengqiang and Yong, Hongwei and Zhang, Lei},
  journal={arXiv preprint arXiv:2401.00877},
  year={2024}
}

@inproceedings{wu2024osediff,
  title={One-Step Effective Diffusion Network for Real-World Image Super-Resolution},
  author={Wu, Rongyuan and Sun, Lingchen and Ma, Zhiyuan and Zhang, Lei},
  booktitle={Advances in Neural Information Processing Systems},
  volume={37},
  pages={92529--92553},
  year={2024}
}

@article{wu2025dp2osr,
  title={DP2O-SR: Direct Perceptual Preference Optimization for Real-World Image Super-Resolution},
  author={Wu, Rongyuan and Sun, Lingchen and Zhang, Zhengqiang and Wang, Shihao and Wu, Tianhe and Yi, Qiaosi and Li, Shuai and Zhang, Lei},
  journal={arXiv preprint arXiv:2510.18851},
  year={2025}
}

@misc{blackforestlabs2025flux2,
  title        = {{FLUX.2}: Next Generation Image Generation},
  author       = {{Black Forest Labs}},
  year         = {2025},
  howpublished = {\url{https://bfl.ai/models/flux-2}},
  note         = {Accessed: 2026-05-01}
}

@misc{blackforestlabs2026flux2dev,
  title        = {{FLUX.2-dev} Model Card},
  author       = {{Black Forest Labs}},
  year         = {2026},
  howpublished = {\url{https://huggingface.co/black-forest-labs/FLUX.2-dev}},
  note         = {Accessed: 2026-05-01}
}

@misc{huggingface2025flux2diffusers,
  title        = {Diffusers Welcomes {FLUX.2}},
  author       = {{Hugging Face}},
  year         = {2025},
  howpublished = {\url{https://huggingface.co/blog/flux-2}},
  note         = {Accessed: 2026-05-01}
}

@inproceedings{rafailov2023dpo,
  title={Direct Preference Optimization: Your Language Model is Secretly a Reward Model},
  author={Rafailov, Rafael and Sharma, Archit and Mitchell, Eric and Manning, Christopher D. and Ermon, Stefano and Finn, Chelsea},
  booktitle={Advances in Neural Information Processing Systems},
  year={2023}
}

@inproceedings{wallace2024diffdpo,
  title={Diffusion Model Alignment Using Direct Preference Optimization},
  author={Wallace, Bram and Dang, Meihua and Rafailov, Rafael and Zhou, Linqi and Lou, Aaron and Purushwalkam, Senthil and Ermon, Stefano and Xiong, Caiming and Joty, Shafiq and Naik, Nikhil},
  booktitle={Proceedings of the IEEE/CVF Conference on Computer Vision and Pattern Recognition},
  pages={8228--8238},
  year={2024}
}

@misc{dong2025tsdsr,
      title={TSD-SR: One-Step Diffusion with Target Score Distillation for Real-World Image Super-Resolution}, 
      author={Linwei Dong and Qingnan Fan and Yihong Guo and Zhonghao Wang and Qi Zhang and Jinwei Chen and Yawei Luo and Changqing Zou},
      year={2025},
      eprint={2411.18263},
      archivePrefix={arXiv},
      primaryClass={cs.CV},
      url={https://arxiv.org/abs/2411.18263}, 
}

@inproceedings{zhang2018lpips,
  title={The Unreasonable Effectiveness of Deep Features as a Perceptual Metric},
  author={Zhang, Richard and Isola, Phillip and Efros, Alexei A. and Shechtman, Eli and Wang, Oliver},
  booktitle={Proceedings of the IEEE Conference on Computer Vision and Pattern Recognition},
  pages={586--595},
  year={2018}
}

@inproceedings{wang2004ssim,
  title={Image Quality Assessment: From Error Visibility to Structural Similarity},
  author={Wang, Zhou and Bovik, Alan C. and Sheikh, Hamid R. and Simoncelli, Eero P.},
  booktitle={IEEE Transactions on Image Processing},
  volume={13},
  number={4},
  pages={600--612},
  year={2004}
}

@inproceedings{ding2020dists,
  title={Image Quality Assessment: Unifying Structure and Texture Similarity},
  author={Ding, Keyan and Ma, Kede and Wang, Shiqi and Simoncelli, Eero P.},
  booktitle={IEEE Transactions on Pattern Analysis and Machine Intelligence},
  year={2020}
}

@inproceedings{ke2021musiq,
  title={MUSIQ: Multi-Scale Image Quality Transformer},
  author={Ke, Junjie and Wang, Qifei and Wang, Yilin and Milanfar, Peyman and Yang, Feng},
  booktitle={Proceedings of the IEEE/CVF International Conference on Computer Vision},
  pages={5148--5157},
  year={2021}
}

@inproceedings{wang2023clipiqa,
  title={Exploring CLIP for Assessing the Look and Feel of Images},
  author={Wang, Jianyi and Chan, Kelvin C. K. and Loy, Chen Change},
  booktitle={Proceedings of the AAAI Conference on Artificial Intelligence},
  volume={37},
  pages={2555--2563},
  year={2023}
}

@inproceedings{yang2022maniqa,
  title={MANIQA: Multi-Dimension Attention Network for No-Reference Image Quality Assessment},
  author={Yang, Sidi and Wu, Tianhe and Shi, Shuwei and Lao, Shanshan and Gong, Yuan and Cao, Mingdeng and Wang, Jiahao and Yang, Yujiu},
  booktitle={Proceedings of the IEEE/CVF Conference on Computer Vision and Pattern Recognition Workshops},
  pages={1191--1200},
  year={2022}
}

@article{mittal2013niqe,
  title={Making a Completely Blind Image Quality Analyzer},
  author={Mittal, Anish and Soundararajan, Rajiv and Bovik, Alan C.},
  journal={IEEE Signal Processing Letters},
  volume={20},
  number={3},
  pages={209--212},
  year={2013}
}

@inproceedings{karras2019ffhq,
  title={A Style-Based Generator Architecture for Generative Adversarial Networks},
  author={Karras, Tero and Laine, Samuli and Aila, Timo},
  booktitle={CVPR},
  year={2019}
}

@inproceedings{agustsson2017ntire,
  title={NTIRE 2017 Challenge on Single Image Super-Resolution: Dataset and Study},
  author={Agustsson, Eirikur and Timofte, Radu},
  booktitle={CVPR Workshops},
  year={2017}
}

@article{lsdir,
  title={LSDIR: A Large Scale Dataset for Image Restoration},
  author={Li, Yawei and others},
  journal={arXiv preprint},
  year={2023}
}

@inproceedings{hu2021lora,
  title     = {LoRA: Low-Rank Adaptation of Large Language Models},
  author    = {Hu, Edward J. and Shen, Yelong and Wallis, Phillip and Allen-Zhu, Zeyuan and Li, Yuanzhi and Wang, Shean and Wang, Lu and Chen, Weizhu},
  booktitle = {International Conference on Learning Representations},
  year      = {2022},
  eprint    = {2106.09685},
  archivePrefix = {arXiv}
}

\small


\newpage
\appendix

\section{Additional Implementation Details}
\label{app:implementation}

We provide additional details about the Flux2SR supervised training setup, the FoA-SR preference optimization procedure, compute resources, and existing assets used in the experiments.

\paragraph{Flux2SR training.}
Flux2SR is trained by fine-tuning LoRA parameters on top of FLUX.2-dev while keeping the base transformer, VAE, and text encoder frozen. Unless otherwise stated, we use rank-64 LoRA, bf16 precision, AdamW optimization, learning rate \(1\times10^{-4}\), weight decay \(1\times10^{-4}\), a constant-with-warmup learning-rate schedule with 200 warmup steps, batch size 1, gradient accumulation 1, and 50K training steps. During sampling, we use 16 sampling steps, guidance 1.0, initial sigma 1.0, sample sigma 1.0, and a cached prompt embedding.

\paragraph{Supervised reconstruction losses.}
The final Flux2SR baseline is trained with latent flow matching and image-space reconstruction losses:
\[
L_{\mathrm{SR}} = L_{\mathrm{FM}} + \lambda_{\mathrm{pix}} L_{\mathrm{charb}} + \lambda_{\mathrm{lpips}} L_{\mathrm{LPIPS}}.
\]
We set \(\lambda_{\mathrm{pix}}=1.0\), use Charbonnier pixel loss, and set \(\lambda_{\mathrm{lpips}}=1.0\). The LPIPS term is linearly warmed up from 0 to its full weight between training steps 2000 and 6000. Pixel reconstruction is active from the beginning of training.

\paragraph{Preference optimization.}
For FoA-SR-Faithful, FoA-SR-Aesthetic, and FoA-SR-Hybrid, we initialize the trainable LoRA adapter from the same Flux2SR checkpoint and use the original Flux2SR adapter as the frozen reference model. We fine-tune only LoRA parameters and keep the FLUX.2 backbone frozen. Preference optimization uses rank-64 LoRA, learning rate \(2\times10^{-5}\), DPO preference strength \(\beta=1000\), batch size 1, gradient accumulation 8, bf16 precision, and 500 optimization steps.

\paragraph{Evaluation protocol.}
All Flux2SR-based methods are evaluated with 16 sampling steps, guidance 1.0, sample sigma 1.0, and the same metric implementation. Final model comparisons use raw, non-normalized metric values. The within-pool normalization used for reward-based candidate mining is not used for test-set evaluation.

\paragraph{Compute resources.}
All model training, candidate generation, and preference optimization experiments were run on a single NVIDIA H200 GPU.

\begin{table}[h]
\centering
\caption{Compute resources used in FoA-SR experiments.}
\label{tab:app_compute}
\small
\setlength{\tabcolsep}{5pt}
\renewcommand{\arraystretch}{1.08}
\begin{tabular}{@{}p{0.43\linewidth}p{0.38\linewidth}p{0.15\linewidth}@{}}
\toprule
Component & Configuration & Approx. time \\
\midrule
Flux2SR supervised training & rank 64, 50K steps, bf16 & 13h 04m \\
Candidate generation, \(M=4\) & 500 inputs, 4 seeds, 16 steps & \(\sim\)5h \\
Additional candidates for \(M=8\) analysis & 500 inputs, extra 4 seeds, 16 steps & \(\sim\)5h \\
Faithful preference training & 500 steps, rank 64 & \(\sim\)4h \\
Aesthetic preference training & 500 steps, rank 64 & \(\sim\)4h \\
Hybrid preference training & 500 steps, rank 64 & \(\sim\)4h \\
IQA evaluation & fixed outputs, same metric code & runtime only \\
\bottomrule
\end{tabular}
\end{table}

\section{Additional Candidate Mining Details}
\label{app:mining}

For each LR input in the 500-image mining set, we generate multiple stochastic candidates from the same Flux2SR baseline by varying only the random seed. The same candidate pool is used to compute Faithful, Aesthetic, and Hybrid rewards. This design ensures that profile disagreement reflects different restoration preferences rather than differences in data, model initialization, or sampling source.

\paragraph{Profile-disagreement statistics.}
Table~\ref{tab:app_profile_disagreement} reports detailed disagreement statistics at the training candidate pool size \(M=4\). Faithful and Aesthetic rewards select different winners for 78.4\% of inputs, and their rank correlation is close to zero, indicating that the two profiles induce different rankings over the same candidate pool.

\begin{table}[h]
\centering
\caption{Detailed profile-disagreement statistics at the training candidate pool size \(M=4\).}
\label{tab:app_profile_disagreement}
\begin{tabular}{lc}
\toprule
Statistic & Value \\
\midrule
Number of LR inputs & 500 \\
Candidates per input & 4 \\
Winner disagreement rate & 78.4\% \\
Mean Spearman \(R_F,R_A\) & -0.0312 \\
Mean Faithful reward gap & 0.6517 \\
Mean Aesthetic reward gap & 0.7097 \\
Mean \(R_F(y_w^F)-R_F(y_w^A)\) & 0.3338 \\
Mean \(R_A(y_w^A)-R_A(y_w^F)\) & 0.3411 \\
\bottomrule
\end{tabular}
\end{table}

\paragraph{Effect of candidate pool size.}
To assess whether the training pool size \(M=4\) provides a sufficient preference-mining signal, we extend the same 500-image mining set up to \(M=8\) candidates and recompute mining statistics for nested pool sizes \(M\in\{2,4,6,8\}\), without retraining any adapter. As shown in Table~\ref{tab:app_candidate_pool_size}, \(M=4\) already provides strong reward gaps and substantial profile disagreement, while increasing the pool size mainly strengthens the same disagreement signal at additional generation and scoring cost.

\begin{table}[h]
\centering
\caption{Detailed candidate-pool-size analysis over 500 LR inputs. We report mean and standard deviation of reward gaps.}
\label{tab:app_candidate_pool_size}
\begin{tabular}{ccccc}
\toprule
\(M\) & Faithful gap & Aesthetic gap & Winner disagreement & Spearman \(R_F,R_A\) \\
\midrule
2 & \(0.5628 \pm 0.3506\) & \(0.5700 \pm 0.3683\) & 55.2\% & -0.1690 \\
4 & \(0.6517 \pm 0.2215\) & \(0.7097 \pm 0.1978\) & 78.4\% & -0.0312 \\
6 & \(0.6749 \pm 0.1848\) & \(0.7216 \pm 0.1598\) & 85.4\% & -0.0429 \\
8 & \(0.6741 \pm 0.1616\) & \(0.7300 \pm 0.1421\) & 90.8\% & -0.0380 \\
\bottomrule
\end{tabular}
\end{table}

\begin{figure}[h]
\centering
\includegraphics[width=\linewidth]{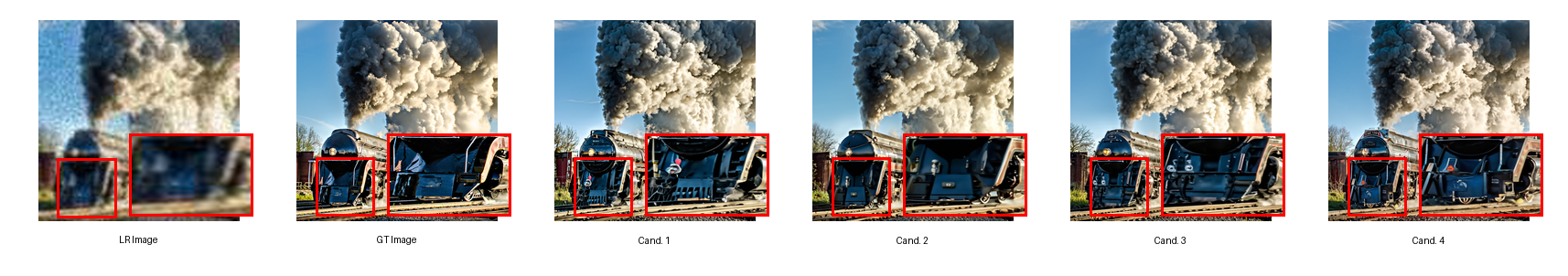}
\caption{Example stochastic candidate pool generated by the shared Flux2SR baseline for the same LR input. The candidates are produced by changing only the random seed. Faithful and Aesthetic rewards may select different winners from the same pool, motivating profile-specific preference mining.}
\label{fig:app_canon_seeds}
\end{figure}

\paragraph{Stochastic candidate examples.}
Figure~\ref{fig:app_canon_seeds} shows a stochastic candidate pool generated by the shared Flux2SR baseline for the same LR input. The candidates differ only by random seed, illustrating that the same baseline can produce multiple plausible restorations.

\section{Extended Quantitative Results}
\label{app:extended_results}

We report additional quantitative results that are not included in the main paper due to space limitations. These include supervised Flux2SR ablations, LoRA target-module ablations, and extended metrics for the Hybrid reward ablation.
\subsection{Supervised Flux2SR Objective Ablation}
\label{app:supervised_ablation}
Table~\ref{tab:app_supervised_objective} compares supervised Flux2SR training objectives on RealSR. FM+Pix obtains the strongest distortion-oriented metrics among the supervised variants, but its learned no-reference metrics are lower. FM+Pix+LPIPS provides a more balanced starting point by improving MUSIQ, MANIQA, and NIQE relative to FM+Pix while retaining reasonable reference fidelity.

\begin{table}[!htbp]
\centering
\caption{Supervised Flux2SR objective ablation on RealSR. All variants use rank-64 LoRA, 50K training steps, the Attn+Add-LoRA target set, and the same 16-step inference setting.}
\label{tab:app_supervised_objective}
\resizebox{\linewidth}{!}{
\begin{tabular}{lccccccccc}
\toprule
Objective & PSNR$\uparrow$ & SSIM$\uparrow$ & LPIPS$\downarrow$ & DISTS$\downarrow$ & CLIPIQA$\uparrow$ & NIQE$\downarrow$ & MUSIQ$\uparrow$ & MANIQA$\uparrow$ & FID$\downarrow$ \\
\midrule
FM only & 22.5933 & 0.6895 & 0.2914 & 0.2182 & 0.6612 & 5.4024 & 67.3828 & 0.6519 & 106.4577 \\
FM + Pix & 24.1125 & 0.7329 & 0.2605 & 0.2009 & 0.5892 & 5.5422 & 65.2518 & 0.6216 & 108.5328 \\
FM + Pix + LPIPS & 22.7195 & 0.6733 & 0.2793 & 0.2191 & 0.6154 & 4.9360 & 69.5302 & 0.6667 & 110.0081 \\
\bottomrule
\end{tabular}
}
\end{table}
We therefore use FM+Pix+LPIPS as the shared Flux2SR baseline for studying both Faithful and Aesthetic preference optimization.

\FloatBarrier

\subsection{LoRA Target-Module Ablation}
\label{app:lora_ablation}

We also ablate which FLUX.2 transformer modules receive LoRA adapters. We compare three target sets. Attn-LoRA applies LoRA to the default attention projection modules. Attn+Add-LoRA further adds LoRA to the additional conditioning projections (\texttt{add\_q\_proj}, \texttt{add\_k\_proj}, \texttt{add\_v\_proj}, and \texttt{to\_add\_out}). Attn+FFN-LoRA instead extends Attn-LoRA with feed-forward modules (\texttt{ff.linear\_in}, \texttt{ff.linear\_out}, \texttt{ff\_context.linear\_in}, and \texttt{ff\_context.linear\_out}).

\begin{table}[!htbp]
\centering
\caption{Ablation of LoRA target modules for supervised Flux2SR training on RealSR. All variants use FM-only training, rank-64 LoRA, 50K steps, and the same 16-step inference setting.}
\label{tab:app_lora_target_ablation}
\resizebox{\linewidth}{!}{
\begin{tabular}{lccccccccc}
\toprule
LoRA target set & PSNR$\uparrow$ & SSIM$\uparrow$ & LPIPS$\downarrow$ & DISTS$\downarrow$ & CLIPIQA$\uparrow$ & NIQE$\downarrow$ & MUSIQ$\uparrow$ & MANIQA$\uparrow$ & FID$\downarrow$ \\
\midrule
Attn-LoRA & 22.1328 & 0.6766 & 0.2957 & 0.2213 & 0.6319 & 5.4714 & 66.1344 & 0.6419 & 106.1353 \\
Attn+Add-LoRA & 22.5933 & 0.6895 & 0.2914 & 0.2182 & 0.6612 & 5.4024 & 67.3828 & 0.6519 & 106.4577 \\
Attn+FFN-LoRA & 21.8387 & 0.6751 & 0.3097 & 0.2310 & 0.6910 & 5.4688 & 69.4892 & 0.6676 & 115.5775 \\
\bottomrule
\end{tabular}
}
\end{table}

Attn+FFN-LoRA improves several learned no-reference metrics, such as CLIPIQA, MUSIQ, and MANIQA, but degrades reference-oriented metrics and FID. Attn+Add-LoRA provides a stronger balance between distortion-oriented reconstruction and learned perceptual quality. We therefore adopt Attn+Add-LoRA as the default target-module set for Flux2SR and the profile-specific adapters.

\FloatBarrier

\subsection{Extended Hybrid Reward Metrics}
\label{app:hybrid_extended}

Table~\ref{tab:app_hybrid_extended} reports additional held-out RealSR metrics for the Hybrid reward ablation.

\begin{table}[!htbp]
\centering
\caption{Extended RealSR metrics for the Hybrid reward ablation. These metrics provide an additional view of how the Hybrid adapter behaves relative to Faithful and Aesthetic endpoints.}
\label{tab:app_hybrid_extended}
\resizebox{\linewidth}{!}{
\begin{tabular}{lcccccccc}
\toprule
Method & TOPIQ-FR$\uparrow$ & AFINE-FR$\downarrow$ & CLIPIQA+$\uparrow$ & TOPIQ-NR$\uparrow$ & AFINE-NR$\downarrow$ & NIMA$\uparrow$ & TOPIQ-IAA$\uparrow$ & FID$\downarrow$ \\
\midrule
Flux2SR & 0.4769 & -0.7410 & 0.6792 & 0.6258 & -1.0478 & 4.8029 & 4.6670 & 110.0081 \\
FoA-SR-Faithful & 0.5146 & -0.9138 & 0.6380 & 0.5847 & -0.9974 & 4.6323 & 4.5808 & 107.1397 \\
FoA-SR-Aesthetic & 0.4118 & -0.4605 & 0.7579 & 0.7408 & -1.0772 & 5.1457 & 5.0062 & 123.7668 \\
FoA-SR-Hybrid & 0.4467 & -0.6020 & 0.7399 & 0.7332 & -1.1005 & 5.0373 & 4.9520 & 121.1101 \\
\bottomrule
\end{tabular}
}
\end{table}

FoA-SR-Hybrid improves many held-out no-reference and aesthetic metrics over Flux2SR, but it does not match FoA-SR-Faithful on held-out reference metrics such as TOPIQ-FR and AFINE-FR, nor does it consistently reach FoA-SR-Aesthetic on the strongest aesthetic metrics. This supports the interpretation that a single averaged reward yields a compromise behavior rather than explicit profile control.

\FloatBarrier

\section{Additional Qualitative Results}
\label{app:qualitative}

\begin{figure}[p]
\centering
\includegraphics[width=\linewidth]{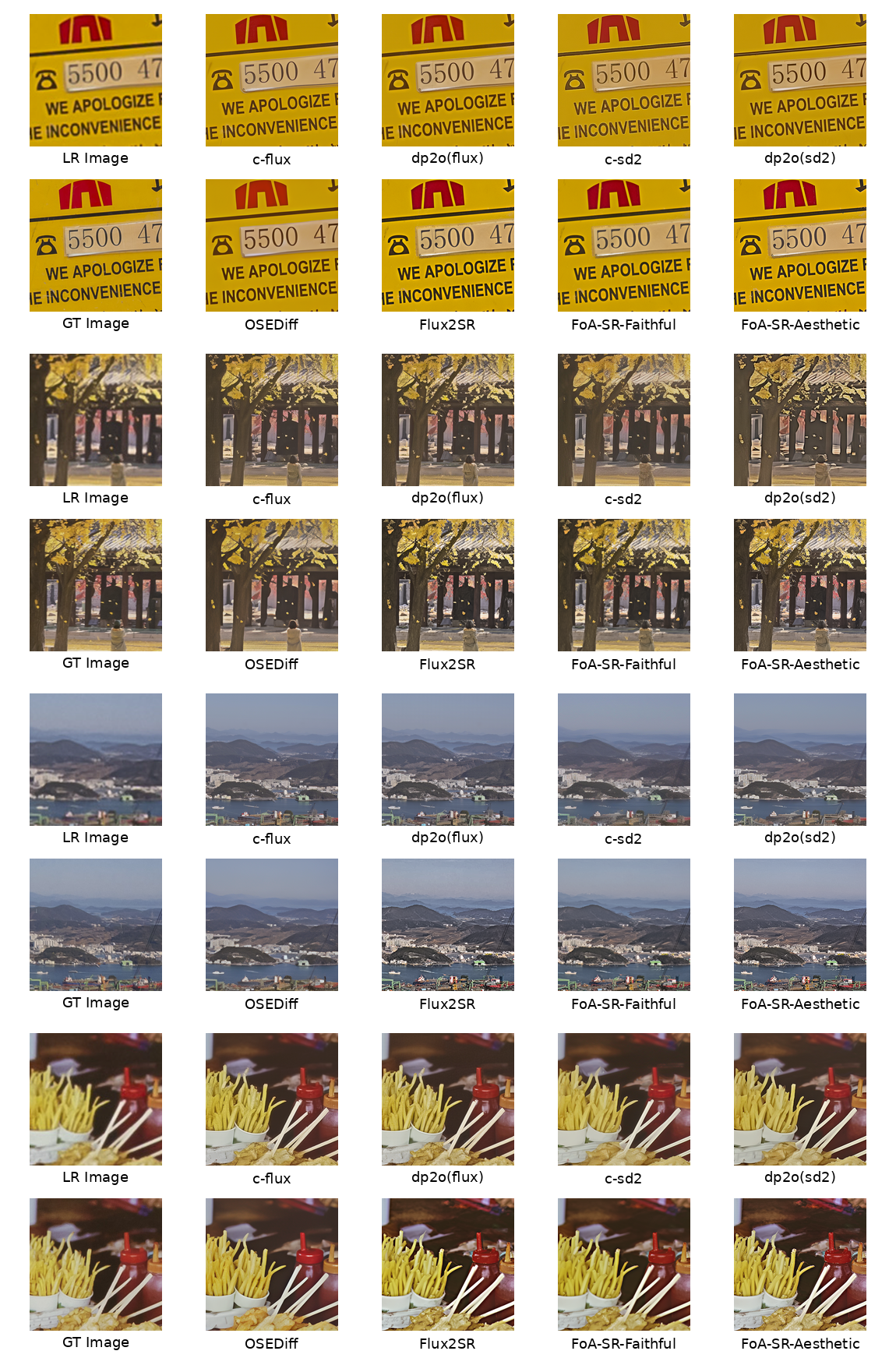}
\caption{Additional qualitative examples on RealSR comparing Flux2SR, FoA-SR-Faithful, FoA-SR-Aesthetic, and other Real-ISR methods. Zoom in for visual details.}
\label{fig:app_qualitative_realsr}
\end{figure}

\begin{figure}[p]
\centering
\includegraphics[width=\linewidth]{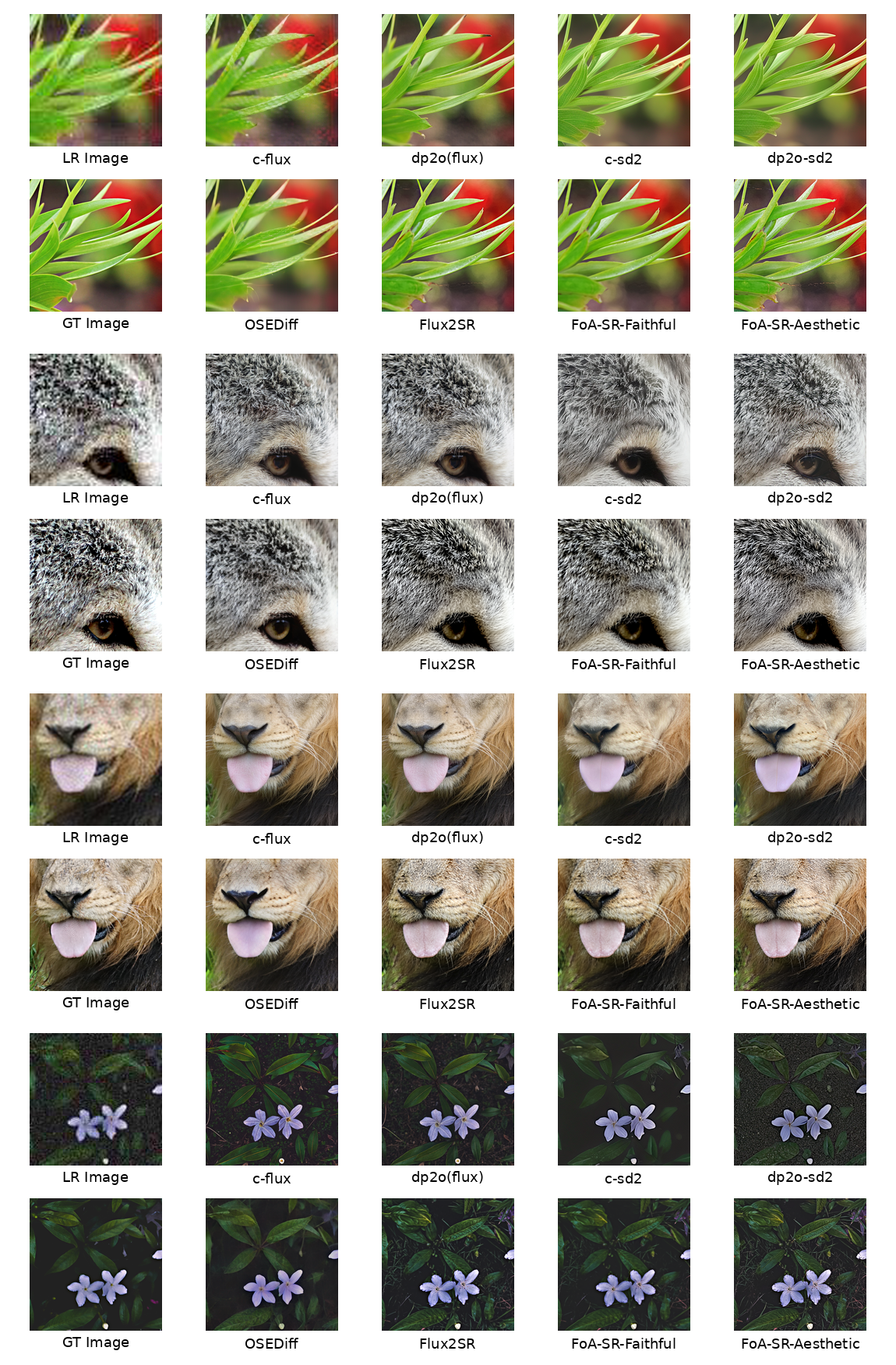}
\caption{Additional qualitative examples on DIV2K-Val comparing Flux2SR, FoA-SR-Faithful, FoA-SR-Aesthetic, and other Real-ISR methods. Zoom in for visual details.}
\label{fig:app_qualitative_div2k}
\end{figure}

Figure~\ref{fig:app_qualitative_realsr} and Figure~\ref{fig:app_qualitative_div2k} show additional qualitative comparisons on RealSR and DIV2K-Val.

\FloatBarrier


\newpage


\end{document}